\documentclass[runningheads]{llncs}
\usepackage[T1]{fontenc}
\usepackage{graphicx}
\usepackage{bbding}

\usepackage{amsmath} 
\usepackage{amsfonts} 
\usepackage{algorithm} 
\usepackage{algorithmic} 

\usepackage{booktabs} 
\usepackage{caption} 
\usepackage{array} 
\usepackage{multirow} 
\usepackage{graphicx} 
\usepackage{xcolor}
\usepackage{subcaption}
\usepackage[colorlinks,
            linkcolor=blue,
            anchorcolor=blue,
            citecolor=blue]{hyperref}

\begin{document}
\title{ Learning from Noisy Labels for Long-tailed Data via Optimal Transport}
%
%
\author{Mengting Li\inst{1}\orcidID{0009-0006-0261-928X} \and
Chuang Zhu\inst{1}\orcidID{0000-0001-5155-7069}\Envelope }
\authorrunning{Mengting Li et al.}
%
\institute{Beijing University of Posts and Telecommunications, Beijing, China \email{\{mtli,czhu\}@bupt.edu.cn}\\
}
\maketitle              
\begin{abstract}
Noisy labels, which are common in real-world datasets, can significantly impair the training of deep learning models. However, recent adversarial noise-combating methods overlook the long-tailed distribution of real data, which can significantly harm the effect of denoising strategies. Meanwhile, the mismanagement of noisy labels further compromises the model's ability to handle long-tailed data. To tackle this issue, we propose a novel approach to manage data characterized by both long-tailed distributions and noisy labels. First, we introduce a loss-distance cross-selection module, which integrates class predictions and feature distributions to filter clean samples, effectively addressing uncertainties introduced by noisy labels and long-tailed distributions. Subsequently, we employ optimal transport strategies to generate pseudo-labels for the noise set in a semi-supervised training manner, enhancing pseudo-label quality while mitigating the effects of sample scarcity caused by the long-tailed distribution. We conduct experiments on both synthetic and real-world datasets, and the comprehensive experimental results demonstrate that our method surpasses current state-of-the-art methods. Our code will be available in the future.

\end{abstract}
\section{Introduction}
In recent years, deep neural networks (DNNs) have gained remarkable achievement in computer vision tasks, such as image classification, object detection, and segmentation. The success of DNNs depends heavily on the availability of rich data. Large datasets like ImageNet \cite{krizhevsky2012imagenet} play an important role in the development of DNNs. However, such abundant and accurately labeled data is not always available in practice. On the one hand, noisy labels are inevitable in real-world datasets from various sources \cite{yu2018learning,xie2019improving,lloyd2004observer}, which can adversely impact the performance of DNNs. Recent evidence suggests that DNNs, with their strong capacity, can easily fit noisy labels during the learning process, resulting in poor generalization performance \cite{zhang2021understanding}. Thus, learning robustly with noisy labels can be a challenging task. It is crucial to process noisy labels for computer vision tasks, especially for classification tasks. On the other hand, class imbalance is also prevalent in real-world settings, with head classes having numerous samples while tail classes are severely underrepresented. This distribution is known as long-tailed distribution \cite{wang2022label,zhong2021improving,hu2020learning}. Noisy label bias and class imbalance often coexist in real-world datasets, significantly impairing model training effectiveness.

Recent research has separately addressed these two issues: learning with noisy labels (LNL) and long-tailed learning (LTL). Current strategies for handling noisy labels include designing noise-robust regularization techniques\cite{liu2020early,tanno2019learning}, developing noise-robust loss functions\cite{ghosh2017robust,lyu2019curriculum,zhou2021learning}, and combining clean sample selection with semi-supervised learning\cite{2020DivideMix,cordeiro2023longremix}. These approaches leverage model memorization effects and the ``small loss'' principle for clean sample identification. However, these strategies may fail in the presence of long-tailed phenomena, as they may misidentify clean samples of minority classes as noisy ones due to their similar high loss values with noisy samples of majority classes \cite{lu2023label}.

For long-tailed learning, most studies employ methods such as re-balancing \cite{shen2016relay,zhou2020bbn}, designing distribution-robust loss functions \cite{cao2019learning,jamal2020rethinking} and decoupling representation learning and classifier learning\cite{kang2019decoupling,zhang2021distribution}. However, these methods overlook noisy labels, which can adversely affect model training by accumulating errors, thereby influencing the long-tailed strategies that rely on model outputs.

It is evident that most of the literature addresses only one of these two problems, with effectiveness compromised when both issues are present simultaneously. Addressing label noise under a long-tailed class distribution remains an ongoing challenge. Recent research\cite{li2023stochastic,zhang2023noisy} focusing on the combined issue typically relies on traditional two-stage noise learning frameworks (sample selection and semi-supervised learning), employing strategies like reassigning weights for clean tail classes, designing loss functions robust to long-tailed distributions and noise, and using sample representations for sample selection strategies. Nevertheless, these approaches overlook the impact of long-tailed noisy data on the quality of pseudo-labels during the semi-supervised learning phase, and filtering strategies during the sample selection phase remain singular.

This article addresses the core issue of noisy-label image classification learning in long-tailed data distributions. To overcome limitations observed in previous studies, we propose the novel OTLNL (Optimal Transport Learning from Noisy Labels) approach. Initially, our loss-distance cross-selection module combines model predictions of sample class with sample feature distributions to filter clean samples effectively, thereby mitigating uncertainties arising from noisy labels and long-tailed distributions using a unified metric. Specifically, we employ dynamic class-specific threshold strategies to filter preliminary clean samples, thereby avoiding the overlap in losses between noisy head class samples and clean tail class samples within long-tailed distributions. Subsequently, utilizing these selected sample features, we compute class centroids, where the distance between class centroids and sample features determines sample cleanliness, reducing the impact of noisy labels on filtering accurate clean and noisy sets.

Next, during the semi-supervised training phase, clean sets are denoted as labeled sets and noisy sets discard labels to form unlabeled sets. We employ optimal transport strategies for generating pseudo-labels for the noisy sets, enhancing pseudo-label quality. Simultaneously, we substitute class sample features with class centroids as optimization targets, mitigating the damage caused by the lack of tail class samples. We are the first to introduce optimal transport strategies for pseudo-label aggregation in the LNL research field. Experimental evaluations on both synthetic and real datasets demonstrate that our method surpasses current state-of-the-art approaches.
The major contributions of this work are as follows:
\begin{enumerate}
    \item[a)] We address the challenging issue of learning from noisy labels in long-tailed data, which is prevalent in real-world scenarios yet often overlooked. 
    \item[b)] We propose a novel approach to handle data with long-tailed distributions and noisy labels, enhancing the accuracy of clean sample selection under long-tailed distributions by introducing dynamic class-specific thresholds in conjunction with feature and label space. 
    \item[c)] We are the first in the LNL research field to design pseudo-label generation optimization using optimal transport strategies. This approach improves training robustness in the presence of noisy labels and class imbalance.
    \item[d)] We conduct experiments on both synthetic and real-world datasets, and outstanding experimental results demonstrate the superiority of our approach over state-of-the-art (SOTA) methods across various datasets and settings.
\end{enumerate}

\section{Related work}

\textbf{Long-tailed learning.} Most studies in LTL utilize three main types of methods: (1) Re-balancing\cite{shen2016relay,zhou2020bbn}: strategies such as resampling to balance the class distribution of the input samples; (2) Designing distribution-robust loss functions\cite{cao2019learning,jamal2020rethinking}: designing loss functions specifically for long-tailed distributions by reweighting classes within the loss function; (3) Decoupling representation learning and classifier learning\cite{kang2019decoupling,zhang2021distribution}: utilizing all samples for representation learning and only the balanced samples for classifier learning. These approaches leverage the rich information of representation training while avoiding the detrimental effects of imbalanced distributions on the classifier. However, these methods generally overlook noisy labels, which can adversely affect model training by accumulating errors, thereby influencing the various long-tailed strategies that rely on model outputs. Consequently, they fail to generalize to noisy datasets.

\textbf{Learning from noisy labels.} Noisy label learning is significant in real-world applications, and recent research primarily falls into three categories: (1) Designing noise-robust regularization techniques\cite{liu2020early,tanno2019learning}: to prevent overfitting during model training by designing regularization techniques that enhance robustness to label noise. (2) Developing noise-robust loss functions\cite{ghosh2017robust,lyu2019curriculum,zhou2021learning}: designing loss functions that are robust to noisy samples to enhance the model's ability to learn from clean samples. (3) Combining clean sample selection with semi-supervised learning\cite{2020DivideMix,cordeiro2023longremix}: these methods typically follow a two-stage framework. The first stage involves selecting clean samples from noisy data, and the second stage uses the clean samples as the labeled set and the noisy samples as the unlabeled set for semi-supervised training. The latest research adopts this two-stage sample selection method. However, these methods rely on the assumption of balanced class distribution, which is unrealistic in real-world scenarios, and the long-tailed distribution problem significantly affects model training effectiveness.

\textbf{Learning with long-tailed noisy data.} To address these two issues in real-world applications simultaneously, recent studies have explored learning from noisy long-tailed data. Initially, simple outlier detection algorithms like LOF are used to select clean samples in work \cite{zhang2023noisy}. For the clean samples, they simulate the Gaussian distribution of each class. The head class is then used to adjust the Gaussian distribution parameters of the tail class, followed by random sampling from the Gaussian distribution for subsequent training. In addition, Cheng et al. \cite{cheng2022instance} design a loss function effective for both class imbalance and noisy labels. ROLT+
 \cite{wei2022robust} propose the ``small distance'' criterion at the feature level to replace the ``small loss'' criterion, using the distance between features and centroids for sample selection in the representation space. SFA \cite{li2023stochastic} uses an auxiliary balanced classifier to address class imbalance issues while maintaining the  ``small distance'' criterion. It employs Bayesian inference to select more robust centroids for sample selection. However, all these studies only utilize a single dimension of predicted probability or representation for sample selection and overlook the impact of long-tailed distribution on the quality of pseudo-label generation for noisy samples during the semi-supervised training phase. To overcome the limitations, we design a method for handling class imbalance and noisy data, employing multi-dimensional criteria for sample selection and optimal transport strategies to improve pseudo-label generation quality, thereby enhancing robustness to real-world data.
\begin{figure}[!t]
\centering
\includegraphics[width=\textwidth]{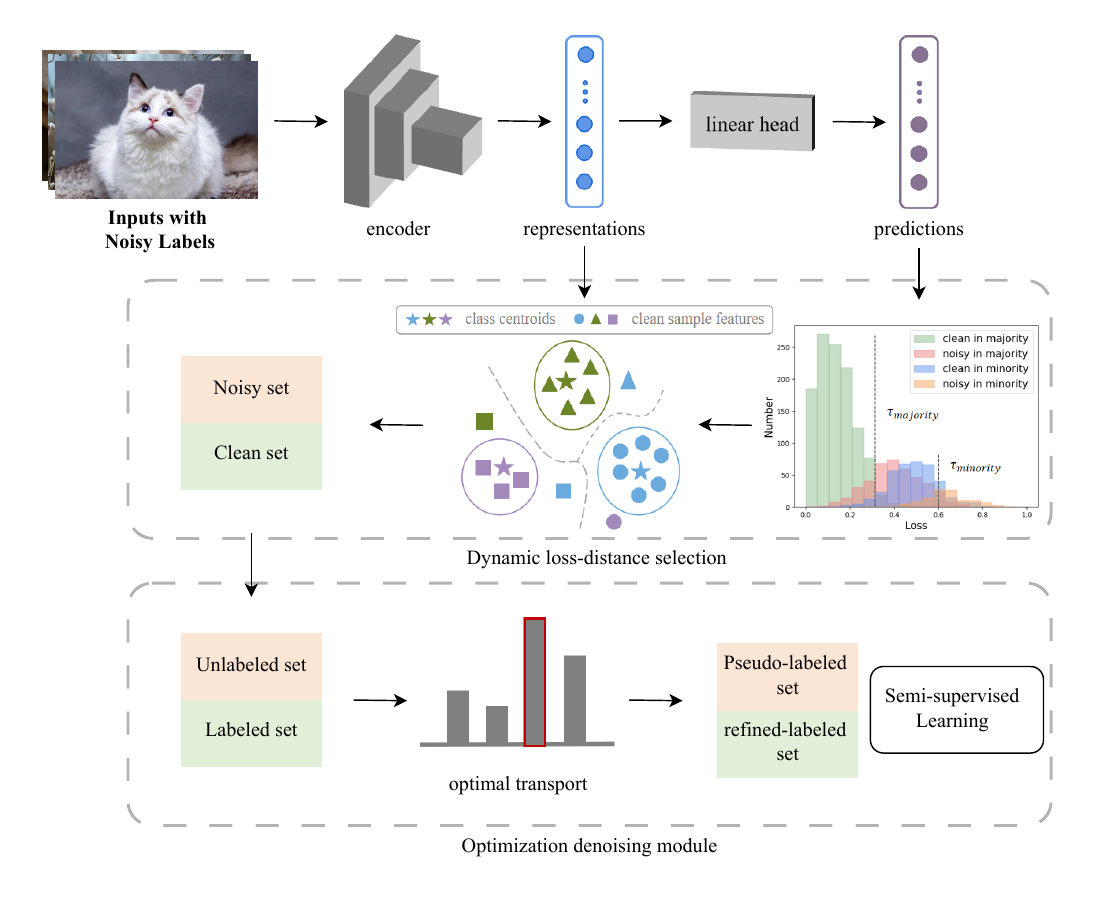}
\caption{The framework of OTLNL. Initially, in the sample selection phase, our loss-distance cross-selection module integrates the model's predictions of sample class probabilities and sample feature distributions to filter clean samples, thereby addressing the uncertainties introduced by noisy labels and long-tailed distributions. Subsequently, during the optimization denoising phase, we employ optimization strategies to generate pseudo-labels for the noise set, enhancing pseudo-label quality while mitigating the effects of sample scarcity.}
\label{fig_1}
\end{figure}
\vspace{-0.8cm}
\vskip 0.4in
\section{Method}
\subsection{Prelimilaries}
This paper considers a typical supervised learning task, a $K$-class image classification problem. Assume that we are provided with a training dataset $D=\{(x_{i},y_{i})\}_{i=1}^N$ of size $N$, where $x_{i}\in X$ is the $i_{th}$ image and $y_{i}\in Y$ is its corresponding label. $X$ denotes the data feature space, and $Y$ denotes the label space over $K$ classes. Specifically, $y$ refers to the image’s observed label. $y$ might be different from the true label because noise can appear in the annotation process in a noisy label scenario. We denote $y^*\in Y^* $as the true label over $K$ classes. Samples of class $k$ are denoted as $D_k$, where $k \in \{1,2,...,K\}$. As this paper is concerned with the problem of long-tailed distribution and noisy labels, we define imbalance ratio as $\rho =  \frac{\max_k |D_k|}{\min_k |D_k|} $ and the fraction of incorrectly labeled training samples as noise ratio $\gamma$. 
\subsection{basic idea}
We propose a novel approach for handling data with long-tailed distributions and noisy labels. Following the two-stage process of previous noise-robust methods, we design a dynamic loss-distance cross-selection module and an optimization denoising module. In the sample selection phase, we integrate the principles of ``small loss'' and ``small distance'' to dynamically filter clean samples in both label and feature spaces, tailored for the characteristics of long-tailed distributions. Subsequently, in the semi-supervised denoising phase, we refine pseudo-label generation for the unlabeled sample set using optimal transport strategies, while employing contrastive loss to enhance feature learning. The overall framework of our approach is illustrated in Fig.~\ref{fig_1}.

\subsection{Dynamic loss-distance selection}
The discovery of the memorization effect \cite{zhang2021understanding} in Deep Neural Networks (DNNs) has had a profound impact on the field of noisy label learning. Specifically, DNNs tend to first memorize easy-to-recognize samples and gradually fit more difficult samples during training. Therefore, we assume that noisy samples tend to output larger losses and are harder to fit in the early stage of training ( the ``warm-up stage'' \cite{2020DivideMix} ). Based on this, the ``small-loss'' criterion has been proposed. Some studies have shown that during training, the losses of clean samples are smaller, while the losses of noisy samples are larger \cite{2020DivideMix,chen2023sample}. Thus, samples with smaller losses are classified as clean, while those with larger losses are classified as noisy. For computational convenience, we use the normalized prediction probability output in place of the loss value, as the loss value is directly calculated from the normalized probability output and the ground truth. That is, samples with higher probabilities are classified as clean.
However, this phenomenon fails in datasets with long-tailed distributions. Research \cite{lu2023label} has found that the loss distribution of clean tail-class samples overlaps with that of noisy head-class samples. Additionally, the loss distribution in tail-class samples is overly concentrated, making it impossible to correctly classify tail-class samples. Consequently, the ``small-distance'' criterion"\cite{li2023stochastic} has been proposed. According to this criterion, the closer a sample's features are to a class centroid, the more likely the sample belongs to that class. The ``small-distance'' criterion relates only to the image itself and not to the labels, thereby reducing the impact of noisy labels. However, the accuracy of centroid calculation is affected by noisy labels and long-tailed distributions.

SFA \cite{li2023stochastic} first utilizes the ``small-distance'' criterion for clean sample selection, where the centroids are calculated using a small portion of clean samples that are selected in the early training stage. However, the selection of this small portion of clean samples is based only on probability outputs and a uniform threshold, which is unreliable and inaccurate. Inspired by this study, we propose a loss-distance dynamic cross-selection method. First, we obtain the prediction probabilities $p(c;x_i)$ in the early training stage ( the ``warm-up stage'' ). Then we generate class-specific dynamic thresholds. Due to the differences in loss (prediction probability) distributions among different classes in long-tailed distributions, the selection thresholds for different classes may vary significantly and should be calculated separately to avoid the loss overlaps. The selection threshold $\tau$ for class $k$ is shown in Formula 1.
\begin{equation}
\tau_{k}(t) = \overline{p}_k(t) = \frac{\sum_{i=1}^{|D_k|} p_{\max}(c; x_i)}{|D_k|},
\end{equation}
where $k \in \{1,2,...,K\}$, and $p_{\max}(c; x_i)$ stands for the maximum prediction probability of sample $x_i$ over $c$ classes.

Additionally, due to the memorization effect, as the number of training epochs increases, the DNN's learning ability improves, and the prediction probability for samples gradually increases. Therefore, the evaluation criterion for sample cleanliness (i.e., the selection threshold) should also increase accordingly, as modified in Formula 2.
\begin{equation}
\tau_k(t) = \lambda \tau_k(t-1) + (1 - \lambda) \tau_k(t),
\end{equation}
where $t$ is the epoch index, and $\lambda$ is the layback ratio that controls selection stability.

To address the issue of overly concentrated loss distribution in tail-class samples, we weight the sample prediction output probabilities used to determine cleanliness. We denote $w(x_i)$ as the weight for modifying prediction probability outputs of $x_i$. A smaller gap between $max(p_i)$ and $p^k_i$ results in a higher weight. To prevent the occurrence of extremely small weights due to division during normalization across all samples in class $k$ for subsequent clustering, we establish a lower bound based on the average prediction confidence of all samples in class $k$, as shown below:
\begin{equation}
w(x_i) = \max \left(p^k_i/\max(p_i), \overline{p_k}^k/\max(\overline{p}_k) \right),
\end{equation}
\begin{equation}
p_i(t) = w(x_i) \times p_i(t),
\end{equation}
where we denote $\overline{p}_k$ as the average prediction confidence of all samples in class $k$.

For samples where the class with the highest prediction probability matches the observed class, we consider them more likely to be clean and assign a larger weight to the prediction probability. Consequently, samples with higher cleanliness probabilities will have higher output probabilities than the original results, making the loss distribution of tail-class samples less compact and easier to classify. With the modified probability outputs, the generated dynamic thresholds apply to both majority and minority classes, mitigating the adverse effects of long-tailed distributions.
After sample selection using dynamic thresholds, the selected small portion of clean samples is used to generate class centroids $\tilde{c}_k$, as shown in Formula 5.
\begin{equation}
\tilde{c}_k = \frac{\sum_{i=1}^{N^l_k} \mathbb{I}\{p_i > \tau_k\} f_{i}}{N^l_k},
\end{equation}
where $\mathbb{I}$ represents the indicator function, $f_i$ stands for the feature representations of $x_i$ and $N^l_k$ is the number of samples that are selected as clean in class $k$.

The calculation of the distance between sample representation and class centroids uses Euclidean distance:

\begin{equation}
\text{dist}(\tilde{c}_k, x_i) = \left\| \tilde{c}_k - f_{i} \right\|_2^2
\end{equation}

The distances are then input into a two-component Gaussian mixture model to determine the clean probability of the sample. Specifically, a sample whose closest centroid class matches the observed class is considered cleaner. At this point, the clean sample set and the noisy sample set have been selected. The ``small-distance'' criterion is based only on image features, reducing the impact of noisy labels.

Additionally, inspired by C2D \cite{zheltonozhskii2022contrast}, we utilize unsupervised contrastive learning for pre-training during the warm-up stage to improve the model's representation ability.

\subsection{Optimization denoising module}
During the semi-supervised denoising phase, the selected clean sample set serves as the labeled set, while the noisy samples are treated as the unlabeled set. Training is conducted using a semi-supervised framework, where the quality of pseudo-labels for the unlabeled set is crucial for the model’s performance. In cross-domain studies, the optimal transport strategy is used to align inter-domain or intra-domain distributions \cite{xu2022semi,huang2023semi}. The objective of this strategy is to find an optimal transport plan for two distributions (given a transport function), which minimizes the transport loss. Inspired by this, in our study, the clean labeled sample set can be treated as the target set, and the unlabeled noisy set as the source set for optimal transport, thereby assigning higher-quality pseudo-labels to the unlabeled noisy set. For each unlabeled sample $x_i$, we create two distinct views using weak and strong augmentations, denoted as $x_i^w$ and $x_i^s$ respectively. For weakly supervised images with labels, we establish a transport plan
\begin{equation}
\gamma_0 = \arg \min_{\gamma} \langle \gamma, C^w \rangle_F,
\end{equation}
where $\gamma_{i,j}$ stands for the transport plan between the i-th labeled clean sample and the j-th unlabeled noisy sample, and $\langle . , . \rangle_F$ represents Frobenius inner product. $C^w$ represents a cost matrix for indicating each transportation.

However, considering the potential long-tailed distribution within the target set, we replace the sample features in the target set with class centroids to mitigate the negative impact of sample scarcity in some classes on the transport plan. Therefore, the transport cost matrix can be expressed as follows: 
\begin{equation}
C^w_{i,j} = 1 - \tilde{c}^k_i \cdot f^w_{j},
\end{equation}
where $\tilde{c}^k_i$ represents the class centroid of the i-th labeled clean sample and $f^w_{j}$ represents the feature representations of the j-th unlabeled noisy sample. 

Minimizing the transport loss yields an optimal transport plan that better aligns the distribution of the pseudo-labeled noisy set with the labeled clean set.

To further improve the quality of pseudo-labels, we incorporate both the model’s predicted pseudo-labels and the optimal transport pseudo-labels into the final pseudo-label selection. Specifically, the pseudo-label generation strategy is applied for the unlabeled set, as follows in Formula 9.
\begin{equation}
\tilde{y}^{uw}_i = 
\begin{cases} 
\arg \max(p^{uw}_i(t)), & p^{uw}_i(t) \geq \tau_1 \\
\gamma_{0i}, & \text{otherwise}
\end{cases},
\end{equation}
where $\tau_1$ is the confidence threshold for using a pseudo-label supplementation strategy.

When the distribution of the noisy set images is close to that of the clean set images, utilizing the optimal transport strategy can better align the distribution of the unlabeled noisy set with the labeled clean set, thereby assigning more accurate and higher-quality pseudo-labels to the unlabeled set. For the strongly augmented version of the same image set, applying the optimal transport plan should yield a mapping closely similar to the weakly augmented set. Therefore, a consistency loss is required to constrain the training : 
\begin{equation}
L_{sw} = \langle \gamma_0, C^s \rangle_F
\end{equation}
Additionally, to enhance the model’s learning capability for the unlabeled set, an unsupervised contrastive loss $L_{con}$ is incorporated. The total loss is
\begin{equation}
L_{total} = L_{SSL} + \lambda_{sw}L_{sw} + \lambda_{c}L_{con},
\end{equation}
where $\lambda_{sw}$ is the consistency loss coefficient and $\lambda_{c}$ is the contrastive loss coefficient. And $L_{SSL}$ stands for the semi-supervised loss.

\section{Experiment}
\subsection{Datasets and implement details}
\textbf{Dataset.} We conduct experiments on three datasets: CIFAR-10, CIFAR-100, and WebVision. For synthetic datasets, following previous works \cite{li2023stochastic}, we introduce noise into the training sets of CIFAR-10 and CIFAR-100 and perform long-tailed resampling. Specifically, generating noisy datasets with long-tailed distributions involves two parameters: the imbalance ratio $\rho$ and the noise ratio $\gamma$. We set the number of samples for the $k$-th class to $N_k = \frac{N}{K\cdot\rho^{\frac{k-1}{K-1}}}$.
The added noise is generated based on the noise transition matrix T, defined as:
\begin{equation}
T_{ij} = \mathbb{P}(Y = j \mid Y^* = i) = 
\begin{cases}
1 - \gamma & \text{if } i = j \\ 
\frac{N_j}{N - N_i} \gamma & \text{otherwise.}
\end{cases}
\end{equation}

For real-world datasets, we choose WebVision, a large-scale dataset containing 1000 classes and 2.4 million images. The training set of WebVision exhibits both long-tailed distribution and noisy labels. For experimental testing, we follow previous work \cite{2020DivideMix} and use only the first 50 classes of the Google subset for our experiments. Additionally, to further validate the effectiveness of various methods in addressing long-tailed issues, we set two additional imbalance ratios by resampling to emphasize the long-tailed distribution, specifically setting $\rho$ to 50 and 100.

\textbf{Implementation Details.}
In the experiments on CIFAR-10, for fair comparisons, all methods use an 18-layer PreAct ResNet \cite{he2016identity} as the backbone and are trained with one NVIDIA GeForce RTX 3090. We train the network using SGD for 200 epochs with a momentum of 0.9, a weight decay of $5 \times 10^{-4}$, and a batch size of 128. The initial learning rate is set as 0.02 and decreases by a factor of 10 after 150 epochs. As previously done in work \cite{2020DivideMix}, three imbalance ratios $\rho$ are adopted: 10, 50 and 100. Two noise ratios are chosen: 20\% and 50\%.
For Webvision, following past experimental settings\cite{li2023stochastic}, we use the Inception-ResNet v2 \cite{szegedy2017inception} architecture and train it using SGD for 100 epochs with a momentum of 0.9, a weight decay of $1 \times 10^{-3}$, a batch size of 32. The initial learning rate is set as 0.01 and decreases by a factor of 10 after 50 epochs. The model is trained with two NVIDIA GeForce RTX 3090.
For hyperparameters, $\tau_1$ is set for 0.7, $\lambda$ is set for 0.99, $\lambda_{sw}$ is set for 0.2 and $\lambda_c$ is set for 0.1.
\subsection{Comparison with State-of-the-Art Methods}
We compare the performance of OTLNL with recent state-of-the-art methods, including three categories: (1) Long-tailed learning (LTL): BBN \cite{zhou2020bbn}, cRT \cite{jamal2020rethinking}; (2) Learning from noisy labels (LNL): ELR+ \cite{liu2020early}, DivideMix \cite{2020DivideMix}; (3) Learning with long-tailed noisy data: MW-Net \cite{shu2019meta}, HAR \cite{cheng2022instance}, RoLT+ \cite{wei2022robust}, and PCL \cite{wei2022robust}. Results for these techniques are from SFA \cite{li2023stochastic}.
Table~\ref{table:test_accuracy} summarizes the experimental results of various methods under different noise and imbalance ratios. We record the best test accuracy over all training epochs and the accuracy at the end of the training. It can be seen that the performance of LNL methods significantly declines under high imbalance settings, indicating that these methods do not account for the adverse effects of long-tailed distributions on denoising strategies. Meanwhile, the performance of LTL methods worsens as the noise ratio increases, indicating an inability to handle the detrimental impact of noisy labels on the training process. However, our framework outperforms other methods in all experimental settings, especially under high noise and high imbalance ratios, due to our class-specific dynamic selection strategy and optimal denoising module. Specifically, in the CIFAR-10 dataset, our method outperforms state-of-the-art methods by 6.53\% at 0.5 noise ratio and 100 imbalance ratio. In the CIFAR-100 dataset, it shows a 3.63\% improvement at the same ratio. In other settings, we can also observe more than a 1.22\% improvement.
\vspace{-0.8cm}
\begin{table}[h!]
\caption{Test accuracy (\%) on synthetic CIFAR datasets with varying levels of noise and imbalance ratios. The best results are in bold.}
    \centering
    \resizebox{\textwidth}{!}{
    \begin{tabular}{c c|c c c c c c|c c c c cc}
        \hline
        \multicolumn{2}{c|}{} & \multicolumn{6}{c|}{CIFAR-10} & \multicolumn{6}{c}{CIFAR-100} \\
        \hline
        \multicolumn{2}{c|}{Noise Level} & \multicolumn{3}{c|}{0.2} & \multicolumn{3}{c|}{0.5} & \multicolumn{3}{c|}{0.2} & \multicolumn{3}{c}{0.5} \\
        \hline
        \multicolumn{2}{c|}{Imbalance Ratio} & 10 & 50 & 100 & 10 & 50 & 100 & 10 & 50 & 100 & 10 & 50 & 100 \\
        \hline
        \multirow{2}{*}{CE} & Best & 77.86 & 64.38 & 61.79 & 60.72 & 46.50 & 38.43 & 45.97 & 33.41 & 29.85 & 28.70 & 18.49 & 16.24 \\
        & Last & 74.00 & 61.38 & 55.69 & 44.29 & 32.55 & 27.28 & 45.75 & 33.12 & 29.58 & 23.70 & 16.56 & 14.19 \\
        \hline
        \multirow{2}{*}{BBN} & Best & 78.44 & 69.05 & 64.24 & 64.51 & 48.88 & 37.75 & 48.60 & 29.08 & 27.44 & 31.05 & 20.33 & 15.51 \\
        & Last & 77.67 & 68.01 & 64.15 & 53.67 & 45.06 & 34.93 & 47.72 & 28.87 & 27.04 & 30.11 & 19.97 & 14.95 \\
        \hline
        \multirow{2}{*}{cRT} & Best & 77.67 & 68.50 & 60.85 & 62.37 & 42.60 & 35.75 & 43.56 & 31.07 & 24.65 & 26.31 & 19.65 & 15.41 \\
        & Last & 75.36 & 67.94 & 58.67 & 60.35 & 41.58 & 33.86 & 42.75 & 30.43 & 23.97 & 25.15 & 19.32 & 14.82 \\
        \hline
        \multirow{2}{*}{ELR+} & Best & 88.96 & 80.21 & 69.60 & 85.02 & 56.96 & 48.72 & 54.01 & 49.64 & 38.40 & 49.53 & 30.12 & 21.58 \\
        & Last & 88.09 & 79.69 & 66.67 & 84.08 & 48.14 & 43.11 & 53.32 & 48.37 & 38.12 & 49.06 & 29.68 & 20.47 \\
        \hline
        \multirow{2}{*}{DivideMix} & Best & 88.79 & 75.34 & 66.90 & 87.54 & 67.92 & 61.81 & 63.79 & 49.64 & 43.91 & 49.35 & 36.52 & 31.82 \\
        & Last & 88.10 & 73.48 & 63.76 & 86.88 & 65.22 & 59.65 & 63.17 & 48.37 & 42.59 & 48.87 & 35.72 & 31.05 \\
        \hline
        \multirow{2}{*}{MW-Net} & Best & 82.19 & 71.63 & 67.26 & 72.12 & 56.09 & 46.36 & 50.20 & 36.68 & 31.77 & 37.50 & 23.99 & 21.24 \\
        & Last & 77.67 & 64.12 & 58.23 & 59.68 & 45.39 & 37.05 & 47.82 & 34.45 & 29.57 & 33.14 & 20.33 & 18.82 \\
        \hline
        \multirow{2}{*}{HAR} & Best & 81.63 & 66.45 & 56.95 & 63.07 & 54.54 & 38.41 & 45.28 & 29.74 & 26.79 & 29.30 & 17.33 & 14.47 \\
        & Last & 78.04 & 60.17 & 54.78 & 61.13 & 48.61 & 35.40 & 44.52 & 26.13 & 23.90 & 26.46 & 14.68 & 12.36 \\
        \hline
        \multirow{2}{*}{RoLT+} & Best & 87.95 & 77.26 & 72.31 & 88.17 & 75.11 & 64.42 & 64.22 & 51.01 & 45.35 & 53.31 & 39.78 & 35.29 \\
        & Last & 87.54 & 75.90 & 69.12 & 87.45 & 73.92 & 61.15 & 63.31 & 49.40 & 43.16 & 52.44 & 39.27 & 34.43 \\
        \hline
        \multirow{2}{*}{PCL} & Best & 90.92 & 84.12 & 79.54 & 84.04 & 71.44 & 66.33 & 65.23 & 51.73 & 47.38 & 57.65 & 42.51 & 38.42 \\
        & Last & 90.81 & 83.71 & 73.84 & 83.51 & 71.44 & 64.69 & 65.14 & 51.46 & 47.12 & 57.65 & 42.51 & 38.36 \\
        \hline
        \multirow{2}{*}{SFA} & Best & 92.53 & 85.96 & 80.26 & 90.57 & 79.89 & 75.17 & 66.32 & 54.29 & 48.51 & 57.41 & 44.47 & 39.73 \\
        & Last & 92.13 & 84.80 & 79.22 & 90.08 & 78.93 & 74.06 & 65.65 & 53.10 & 47.73 & 57.28 & 43.41 & 39.73 \\
        \hline
        \multirow{2}{*}{OTLNL (ours)} & Best & \textbf{93.75} & \textbf{87.94} & \textbf{84.15} & \textbf{91.24} & \textbf{84.98} & \textbf{81.70} & \textbf{67.91} & \textbf{57.12} & \textbf{52.17} & \textbf{62.15} & \textbf{49.61} & \textbf{43.36} \\
        & Last & 93.25 & 87.06 & 84.15 & 90.86 & 84.91 & 81.08 & 67.36 & 56.80 & 51.95 & 61.07 & 49.61 & 43.15 \\
        \hline
    \end{tabular}
    }
    \label{table:test_accuracy}
\end{table}
\vskip 0.4in
Table~\ref{table:webvision} shows the experimental results on the WebVision dataset. It can be seen that our method achieves superior results under high imbalance ratios, demonstrating its effective optimization for training models on data with real-world noise and long-tailed distributions. Specifically, for Webvision under the $\rho=50$ setting, we achieve a top-1 accuracy of 71.43\%, surpassing the state-of-the-art method by 0.79\%. Under the $\rho=100$ setting, we achieve a top-1 accuracy of 66.55\%, surpassing the state-of-the-art methods by 0.87\%. Even on the original dataset with $\rho=6$, without additional long-tailed resampling, our accuracy is still comparable to the state-of-the-art.
\vspace{-0.8cm}
\begin{table}[h]
\caption{Test accuracy (\%) on mini-WebVision and ImageNet with various imbalance ratios.}
\centering
\begin{tabular}{c l c c c c}
\hline
\textbf{IR} & \textbf{Method} & \multicolumn{2}{c}{\textbf{WebVision}} & \multicolumn{2}{c}{\textbf{ImageNet}} \\ \cline{3-6}
 &  & \textbf{top1} & \textbf{top5} & \textbf{top1} & \textbf{top5} \\ \hline
$\rho \approx 6$ & ELR+ & 77.78 & 91.68 & 70.29 & 89.76 \\ 
 & DivideMix & 77.32 & 91.64 & 75.20 & 90.84 \\ 
 & HAR & 75.50 & 90.70 & 70.30 & 90.00 \\ 
 & RoLT+ & 77.64 & 92.44 & 74.64 & 92.48 \\ 
 & PCL & 77.32 & 92.60 & 75.12 & 91.92 \\
 & SFA & 78.96 & \textbf{93.00} & 76.16 & 92.68 \\ 
 & OTLNL (ours) & \textbf{79.12} & 92.56 & \textbf{76.26} & \textbf{93.01} \\ \hline
$\rho = 50$ & DivideMix & 64.56 & 83.56 & 62.68 & 85.24 \\ 
 & RoLT+ & 66.28 & 88.68 & 64.76 & 89.96 \\ 
 & PCL & 68.00 & 88.44 & 65.00 & 86.32 \\ 
 & SFA & 70.64 & 89.96 & \textbf{69.04} & 90.36 \\
 & OTLNL (ours) & \textbf{71.43} & \textbf{89.87} & 68.41 & \textbf{90.43} \\ \hline
$\rho = 100$ & DivideMix & 55.76 & 73.48 & 53.92 & 74.00 \\ 
 & RoLT+ & 60.68 & 87.84 & 59.68 & 88.52 \\ 
 & PCL & 62.12 & 85.88 & 59.60 & 84.20 \\ 
 & SFA & 65.68 & 88.52 & 65.08 & \textbf{88.92} \\ 
 & OTLNL (ours) & \textbf{66.55} & \textbf{88.64} & \textbf{65.26} & 88.64 \\ \hline
\end{tabular}
\label{table:webvision}
\end{table}
\vspace{-0.8cm}
\subsection{Ablation Study and Further Analysis}
\textbf{Ablation Study.} We verify the contribution of each module to the success of our method by removing key modules, with the results presented in Table~\ref{table:ablation}.
For the class-specific dynamic threshold module(CDT), we replace the dynamic threshold strategy with a fixed unified threshold for selecting clean samples for centroid calculation. The results indicate that the dynamic threshold strategy effectively selects a highly pure small clean set, aiding accurate centroid calculation and enhancing the final clean and noisy set selection. The replacement of the threshold results in a 1.72\% drop in test accuracy under 0.5 noise ratio and 100 imbalance ratio.

For the loss-distance cross-selection module(LCS), we replace the entire loss-distance cross-selection module with a time-varying unified threshold. The experimental results highlight the importance of the class dynamic threshold for tail class sample selection and the effectiveness of combined filtering of noisy samples in the feature and label spaces, avoiding incorrect selection for tail classes caused by a single metric. Removing this module results in a 2.25\% drop in test accuracy.

\begin{figure}[ht]
    \centering
    \begin{subfigure}[b]{0.49\textwidth}
        \centering
        \includegraphics[width=\textwidth]{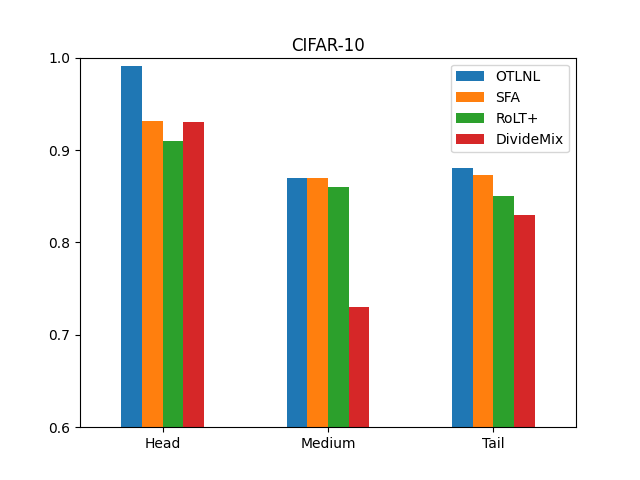}
        \label{fig:image1}
    \end{subfigure}
    \hfill
    \begin{subfigure}[b]{0.49\textwidth}
        \centering
        \includegraphics[width=\textwidth]{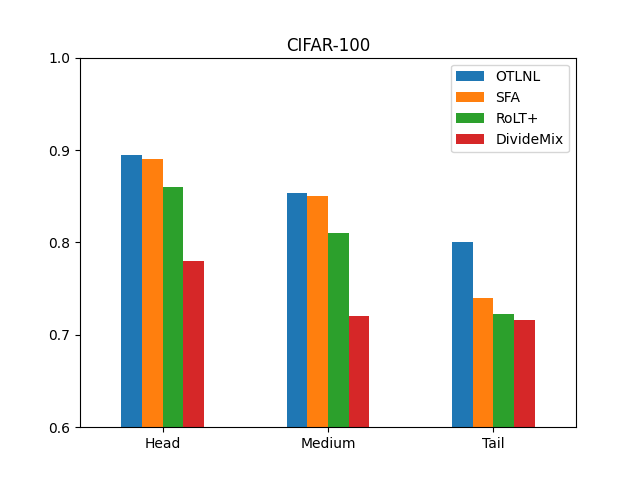}
        \label{fig:image2}
    \end{subfigure}
    \caption{F1-score of sample selection for the head, medium and tail classes on CIFAR-10 and CIFAR-100 datasets under $\gamma = 0.5$ and $\rho = 100$}
    \label{fig:both_images}
\end{figure}

For the optimal denoising module, we remove the optimal transport strategy for pseudo-label prediction(OTP), using only the model-predicted probability output for pseudo-label generation. The table shows that the quality of pseudo-labels significantly impacts the experimental results, with the optimal transport strategy playing a crucial role in refining pseudo-labels for the noisy set during semi-supervised training. The reduction in accuracy indicates that the optimal transport strategy of pseudo-labels provides a 4.15\% improvement for our method.
In summary, each of the key modules mentioned above contributes to an increase in accuracy, but the optimal denoising module plays the most significant role, providing the greatest performance enhancement.
\vspace{-0.8cm}
\begin{table}[h!]
\caption{Ablation studies on key components of our proposed OTLNL framework. We report the test accuracy on CIFAR-10 dataset.}
    \centering
    \begin{tabular}{c c|c c c|c c c}
        \hline
        \multicolumn{2}{c|}{Noise Level} & \multicolumn{3}{c|}{0.2} & \multicolumn{3}{c}{0.5} \\
        \hline
        \multicolumn{2}{c|}{Imbalance Ratio} & 10 & 50 & 100 & 10 & 50 & 100 \\
        \hline
        \multirow{2}{*}{OTLNL} & Best & \textbf{93.75} & \textbf{87.94} & \textbf{84.15} & \textbf{91.24} & \textbf{84.98} & \textbf{81.70} \\
        & Last & 93.25 & 87.06 & 84.15 & 90.86 & 84.91 & 81.08  \\
        \hline
        \multirow{2}{*}{w/o CDT} & Best & 93.01 & 87.05 & 82.63 & 90.70 & 83.88 & 79.98 \\
        & Last & 92.73 & 86.14 & 82.07 & 90.42 & 82.71 & 79.81 \\
        \hline
        \multirow{2}{*}{w/o LCS} & Best & 93.12 & 86.89 & 82.34 & 90.56 & 83.01 & 79.45 \\
        & Last & 93.02 & 86.55 & 82.27 & 90.33 & 82.88 & 79.23 \\
        \hline
        \multirow{2}{*}{w/o OTP} & Best & 91.84 & 83.88 & 80.03 & 89.95 & 82.47 & 77.55 \\
        & Last & 91.59 & 83.51 & 79.77 & 89.69 & 82.19 & 77.04 \\
        \hline
    \end{tabular}
    \label{table:ablation}
\end{table}
\vspace{-0.8cm}
\vskip 0.2in
\textbf{Sample Selection Efficiency.} We analyze the F1-scores of head, middle, and tail classes during the sample selection phase and compare them with representative two-stage sample selection methods ( SFA, RoLT+, DivideMix ). The experimental results in Fig.~\ref{fig:both_images} demonstrate that our method significantly outperforms others in terms of sample selection efficiency, particularly for tail class samples. Our method greatly enhances the learning of tail class samples during model training. The reasons for these results are as follows: (1) We combine the ``small loss'' and ``small distance'' criteria, jointly selecting samples in both label and feature spaces, thereby avoiding the biases of individual criteria; (2) We improve the ``small loss'' strategy by setting thresholds for each class and dynamically increasing them as training epochs increase. This approach aligns better with the theory that a model's learning capacity gradually improves over time, addressing the issue of tail sample training effectiveness being compromised by head class samples.
\vspace{-0.8cm}
\vspace{-0.8cm}
\vskip 0.7in
\section{Conclusion}
To address the concurrent challenges of noisy labels and long-tailed distributions, our work introduces an innovative method rooted in optimal transport theory. Firstly, we propose a dynamic loss-distance cross-selection module, for the identification and filtering of noisy samples. Tailored to accommodate the characteristics of long-tailed distributions, this module dynamically adjusts thresholds for each class by integrating sample prediction probabilities with feature representations, thereby enhancing the robustness of sample selection across all classes. Subsequently, an optimal transport strategy is employed for refining pseudo-labels. A critical innovation here involves substituting class centroids for individual sample features in the target plan, effectively mitigating the detrimental effects of sample scarcity in tail classes on the transport strategy. To our knowledge, we are the first to apply optimal transport theory within the domain of noisy labels, significantly improving the quality of pseudo-labels in noisy datasets. Extensive experiments on both synthetic and real-world datasets demonstrate the superiority of our proposed method over state-of-the-art (SOTA) approaches. Our results not only validate the efficacy of integrating optimal transport theory into noisy label learning, but also highlight its ability to enhance model performance in scenarios characterized by both noisy labels and long-tailed distributions. 
\begin{credits}
\subsubsection{\ackname}This work was supported by the National Key R\&D Program of China (2021ZD0109802), and by the High-performance Computing Platform of BUPT.
\end{credits}
%
%
%
\bibliography{main}

\end{document}